# FACE RECOGNITION SYSTEM

BY:

Yang Li

YLi@my.harrisburgu.edu

Sangwhan Cha, PhD.

Assistant Professor of Computer Science

scha@harrisburgu.edu

# Table of Contents

# Abstract

How to accurately and effectively identify people has always been an interesting topic, both in research and in industry. With the rapid development of artificial intelligence in recent years, facial recognition gains more and more attention. Compared with the traditional card recognition, fingerprint recognition and iris recognition, face recognition has many advantages, including but limit to non-contact, high concurrency, and user friendly. It has high potential to be used in government, public facilities, security, e-commerce, retailing, education and many other fields.

Deep learning is one of the new and important branches in machine learning. Deep learning refers to a set of algorithms that solve various problems such as images and texts by using various machine learning algorithms in multi-layer neural networks. Deep learning can be classified as a neural network from the general category, but there are many changes in the concrete realization. At the core of deep learning is feature learning, which is designed to obtain hierarchical information through hierarchical networks, so as to solve the important problems that previously required artificial design features. Deep Learning is a framework that contains several important algorithms. For different applications (images, voice, text), you need to use different network models to achieve better results.

With the development of deep learning and the introduction of deep convolutional neural networks, the accuracy and speed of face recognition have made great strides. However, as we said above, the results from different networks and models are very different. In this paper, facial features are extracted by merging and comparing multiple models, and then a deep neural network is constructed to train and construct the combined features. In this way, the advantages of multiple models can be combined to mention the recognition accuracy. After getting a model with high accuracy, we build a product model. This article compares the pure-client model with the server-client model, analyzes the pros and cons of the two models, and analyzes the various commercial products that are required for the server-client model.

# Chapter 1

# Face Recognition System

## 1.1 INTRODUCTION

Ever since IBM introduced first personal computer on 1981, to the .com era in the early 2000s, to the online shopping trend in last 10 years, and the Internet of Things today, computers and information technologies are rapidly integrating into everyday human life. As the digital world and real world merge more and more together, how to accurately and effectively identify users and improve information security has become an important research topic.

Not only in the civil area, in particular, since the 9-11 terrorist attacks, governments all over the world have made urgent demands on this issue, prompting the development of emerging identification methods. Traditional identity recognition technology mainly rely on the individual's own memory (password, username, etc.) or foreign objects (ID card, key, etc.). However, whether by virtue of foreign objects or their own memory, there are serious security risks. It is not only difficult to regain the original identity material, but also the identity information is easily acquired by others if the identification items that prove their identity are stolen or forgotten. As a result, if the identity is impersonated by others, then there will be serious consequences.

Different from the traditional identity recognition technology, biometrics is the use of the inherent characteristics of the body for identification, such as fingerprints, irises, face and so on.

Compared with the traditional identity recognition technology, biological features have many advantages, as: 1. Reproducibility, biological characteristics are born with, cannot be changed, so it is impossible to copy other people's biological characteristics. 2. Availability, biological features as part of the human body, readily available, and will never be forgotten. 3. Easy to use. Many biological characteristics will not require individuals to corporate with the examine device. Based on the above advantages, biometrics has attracted the attention of major corporations and research institutes and has successfully replaced traditional recognition technologies in many fields. And with the rapid development of computer and artificial intelligence, biometrics technology is easy to cooperate with computers and networks to realize automation management, and is rapidly integrating into people's daily life.

When comparing the differences between different biometrics, we can see that the cost of facial recognition is low, the acceptance from user is easy, and the acquisition of information is easy. Facial recognition is the use of computer vision technology and related algorithms, from the pictures or videos to find faces, and then analysis of the identity. In addition, further analysis of the acquired face, may conduct some additional attributes of the individual, such as gender, age, emotion, and etc.

## 1.2 APPLICATION OF THIS RESEARCH

Face recognition can be traced back to the sixties and seventies of the last century, and after decades of twists and turns of development has matured. The traditional face detection method relies mainly on the structural features of the face and the color characteristics of the face.

Some traditional face recognition algorithms identify facial features by extracting landmarks, or features, from an image of the subject's face. For example, as shown in Figure 1.1, an algorithm may analyze the relative position, size, and/or shape of the eyes, nose, cheekbones, and jaw. These features are then used to search for other images with matching features. These kinds of algorithms can be complicated, require lots of compute power, hence could be slow in performance. And they can also be inaccurate when the faces show clear emotional expressions, since the size and position of the landmarks can be altered significantly in such circumstance.

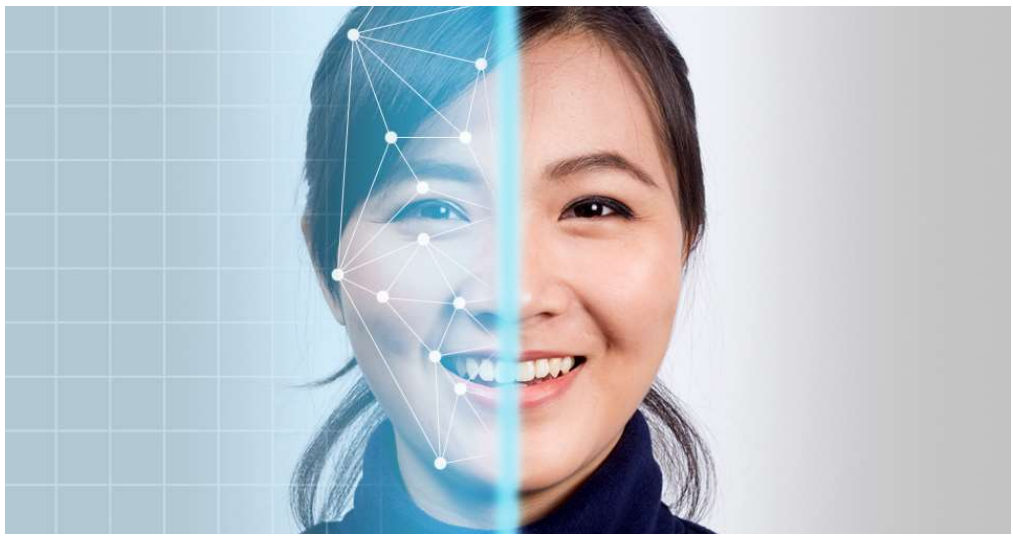

Figure 1.1 Abstract humane face into features [1]

## 1.3 ANALYSIS OF PROBLEM STATEMENT

A complete face recognition system includes face detection, face preprocessing and face recognition processes. Therefore, it is necessary to extract the face region from the face detection process and separate the face from the background pattern, which provides the basis for the subsequent extraction of the face difference features. The recent rise of the face based on the depth of learning detection methods, compared to the traditional method not only shorten the time, and the accuracy is effectively improved. Face recognition of the separated faces is a process of feature extraction and contrast identification of the normalized face images in order to obtain the identity of human faces in the images.

In this paper, we will first summarize and analyze the present research results of face recognition technology, and studies a face recognition algorithm based on feature fusion. The algorithm flow consists of face image preprocessing, combination feature construction and combination feature training.

# Chapter 2

# Theoretical Background

## 2.1 ANALYSIS OF RELATED WORK

In Chapter 1, we introduced the facial recognition, discussed the use case and bright future of this technology. A tremendous amount of research and effort from many major company and universities and been dedicated to this field. In the first part of this chapter, we will review the most significant work in the facial recognition field.

## 2.1.1 FACE DETECTION AND FACE TRACKING

This article *Robust Real-time Object Detection* [2] is the most frequently cited article in a series of articles by Viola that makes face detection truly workable. We can learn about several face detection methods and algorithm from this publication. The article *Fast rotation invariant multi-view face detection based on real Adaboost* [3] for the first time real adaboost applied to object detection, and proposed a more mature and practical multi-face detection framework, the nest structure mentioned on the cascade structure improvements also have good results. The article *Tracking in Low Frame Rate Video: A Cascade Particle Filter with Discriminative Observers of Different Life Spans* [4] is a good combination of face detection model and tracking, offline model and online model, and obtained the CVPR 2007 Best Student Paper.

The above 3 papers discussed about the face detection and face tracking problems. According to the research result in these papers, we can make real time face detection systems. The main purpose is to find the position and size of each face in the image or video, but for tracking, it is also necessary to determine the correspondence between different faces in the frame.

## 2.1.2 FACE POSITIONING AND ALIGNMENT

Earlier localization of facial feature points focused on two or three key points, such as locating the center of the eyeball and the center of the mouth, but later introduced more points and added mutual restraint to improve the accuracy and stability of positioning Sex. The article *Active Shape Models-Their Training and Application* [5] is a model of dozens of facial feature points and texture and positional relationship constraints considered together for calculation. Although ASM has more articles to improve, it is worth mentioning that the AMM model, but also another important idea is to improve the original article based on the edge of the texture model. The regression-based approach presented in the paper *Boosted Regression Active Shape Models* [6] is better than the one based on the categorical apparent model. The article *Face Alignment by Explicit Shape Regression* [7] is another aspect of ASM improvement and an improvement on the shape model itself. Is based on the linear combination of training samples to constrain the shape, the effect of alignment is currently seen the best.

The purpose of the facial feature point positioning is to further determine facial feature points (eyes, mouth center points, eyes, mouth contour points, organ contour points, etc.) on the basis of the face area detected by the face detection / tracking, s position. These 3 articles show the methods for face positioning and face alignment. The basic idea of locating the face feature points is to combine the texture features of the face locals and the position constraints of the organ feature points.

### 2.1.3   FACE FEATURE EXTRACTION

PCA-based eigenfaces [8] are one of the most classic algorithms for face recognition. Although today's PCA is more used in dimensionality reduction in real systems than classification, such a classic approach deserves our attention. The article *Local Gabor Binary Pattern Histogram Sequence (LGBPHS): A Novel Non-Statistical Model for Face Representation and Recognition* [9] is close to many mature commercial systems. In many practical systems, a framework for extracting authentication information is PCA and LDA. Using PDA to reduce matrix to avoid the matrix singularity problem of LDA solving, then using LDA to extract the features suitable for classification, To further identify the various original features extracted after the decision-level fusion. Although some of the LFW test protocols are not reasonable, there is indeed a face recognition library that is closest to the actual data. In this article, *Blessing Dimensionality: High-dimensional Feature and Its Efficient Compression for Face Verification* [10], the use of precise positioning point as a reference to face multi-scale, multi-regional representation of the idea is worth learning, can be combined with a variety of representation.

The above 3 papers discussed about facial feature positioning/alignment. Facial feature extraction is a face image into a string of fixed-length numerical process. This string of numbers is called the "Face Feature" and has the ability to characterize this face. Human face to mention the characteristics of the process of input is "a face map" and "facial features key points coordinates", the output is the corresponding face of a numerical string (feature). Face to face feature algorithm will be based on facial features of the key point coordinates of the human face pre-determined mode, and then calculate the features. In recent years, the deep learning method basically ruled the face lift feature algorithm, In the articles mentioned above, they showed the progress of research in this area. These algorithms are fixed time length algorithm. Earlier face feature models are larger, slow, only used in the background service. However, some recent studies can optimize the model size and operation speed to be available to the mobile terminal under the premise of the basic guarantee algorithm effect.

## 2.2 THEORETICAL IDEA OF PROPOSED WORK

Face recognition is essentially pattern recognition, and the purpose is to abstract real things into numbers that computers can understand. If a picture is a 256 bit-color image, then each pixel of the image is a value between 0 and 255, so we can convert an image into a matrix. How to identify the patterns in this matrix? One way is to use a relatively small matrix to sweep from left to right and top to bottom in this large matrix. Within each small matrix block, we can count the number of occurrences of each color from 0 to 255. So we can express the characteristics of this block.

Through this scan, we get another matrix consisting of many small matrix block features. And this matrix is smaller than the original matrix. Then, for this smaller matrix, perform the above steps again to perform a feature "concentration". In another sense, it is abstracted. Finally, after many abstractions, we will turn the original matrix into a 1 dimension by 1 dimension matrix, which is a number. Different pictures, such as a cat, or a dog, a bear, will eventually get abstracted to different numbers. Similarly, faces, expressions, ages, these principles are similar, but the initial sample size will be large, and ultimately the specific image is abstracted into numbers through the matrix. Then by calculating the difference between the matrixes, we can achieve the goal of comparing faces.

# Chapter 3

# Building Face Recognition Model with Neural Network

## 3.1 INTRODUCTION TO NEURAL NETWORK

Artificial Neural Network (ANN) is a research hotspot in the field of artificial intelligence since the 1980s. It abstracts the human brain neuron network from the perspective of information processing, establishes a simple model, and forms different networks according to different connection methods. It is also often referred to as neural network or neural network in engineering and academia. A neural network is an operational model consisting of a large number of nodes (or neurons) connected to each other. Each node represents a specific output function called an activation function. The connection between every two nodes represents a weighting value for passing the connection signal, called weight, which is equivalent to the memory of the artificial neural network. The output of the network varies depending on the connection method of the network, the weight value and the excitation function. The network itself is usually an approximation of an algorithm or function in nature, or it may be an expression of a logic strategy.

In the past ten years, the research work of artificial neural networks has been deepened, and great progress has been made. It has successfully solved many problems in the fields of pattern recognition, intelligent robots, automatic control, predictive estimation, biology, medicine, and economy. Practical problems that are difficult to solve in modern computers, showing good intelligence.

The artificial neural network model mainly considers the topology of the network connection, the characteristics of the neurons, and the learning rules. At present, there are nearly 40 kinds of neural network models, including back propagation network, perceptron, self-organizing map, Hopfield network, Boltzmann machine, adaptive resonance theory and so on. According to the topology of the connection, the neural network model can be divided into: Feedforward network and Feedback network.

Feedforward network: Each neuron in the network accepts the input of the previous stage and outputs it to the next stage. There is no feedback in the network, and it can be represented by a directed loop-free graph. This kind of network realizes the transformation of signals from input space to output space, and its information processing capability comes from multiple recombination of simple nonlinear functions. The network structure is simple and easy to implement. The backhaul network is a typical forward network.

Feedback network: There is feedback between neurons in the network, which can be represented by an undirected complete graph. The information processing of this neural network is a state transition that can be handled by dynamic system theory. The stability of the system is closely related to the associative memory function. Both the Hopfield network and the Boltzmann machine belong to this type.

## 3.2 CONVOLUTIONAL NEURAL NETWORK

Convolutional neural network (CNN) is a deformation of multi-layer perceptron inspired by biological vision and the most simplified preprocessing operation. It is essentially a forward feedback neural network. The biggest difference between convolutional neural network and multi-layer perceptron is network. The first few layers are composed of a convolutional layer and a pooled layer alternately cascaded to simulate a simple cascade of cells and complex cells for high-level feature extraction in the visual cortex.

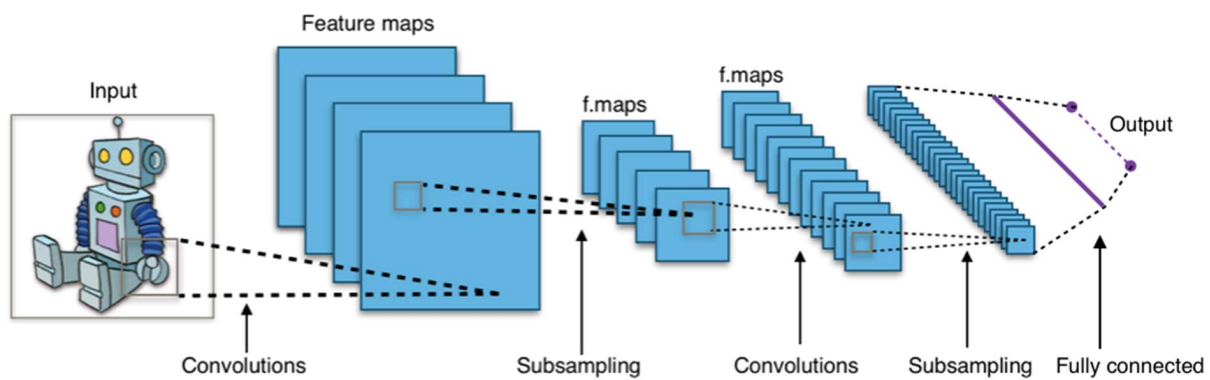

Figure 2. Typical convolutional neural network (CNN) structure [11]

The convolutional neurons respond to a portion of the input from the previous layer (called the local receptive field, with overlap between the regions), extracting higher-level features of the input; the neurons of the pooled layer are input to the previous layer. A portion of the area (no overlap between the areas) is averaged or maximized to resist slight deformation or displacement of the input. The latter layers of the convolutional neural network are typically an output layer of a number of fully connected layers and a classifier.

## 3.3 BUILD FACE RECOGNITION MODEL WITH CNN

At present, face recognition algorithms can be roughly divided into two categories:

(1) Representation-based methods. The basic idea is to convert two-dimensional face input into another space, and then use statistical methods to analyze face patterns, such as Eigenface, Fisherface, and SVM.

(2) A feature-based method generally extracts local or global features and then sends a classifier for face recognition, such as recognition based on set features and HMM.

Convolutional neural network for face recognition can be considered as a feature-based method. It is different from traditional artificial feature extraction and high-performance classifier design for features. Its advantage is that feature extraction is performed by layer-by-layer convolution dimension reduction, and then through multi-layer nonlinear mapping, the network can automatically learn from the unprocessed training samples to form a feature extractor and classifier that adapts to the recognition task. This method reduces the requirements on the training samples, and the number of layers of the network. The more it learns, the more global the features are.

### 3.3.1 THEORY

Convolutional neural network is a deformation of multi-layer perceptron inspired by biological vision and the most simplified preprocessing operation. It is essentially a forward feedback neural network. The biggest difference between convolutional neural network and multi-layer perceptron is network. The first few layers are composed of a convolutional layer and a pooled layer alternately cascaded to simulate a simple cascade of cells and complex cells for high-level feature extraction in the visual cortex.

The convolutional neurons respond to a portion of the input from the previous layer (called the local receptive field, with overlap between the regions), extracting higher-level features of the input; the neurons of the pooled layer are input to the previous layer. A portion of the area (no overlap between the areas) is averaged or maximized to resist slight deformation or displacement of the input. The latter layers of the convolutional neural network are typically an output layer of a number of fully connected layers and a classifier.

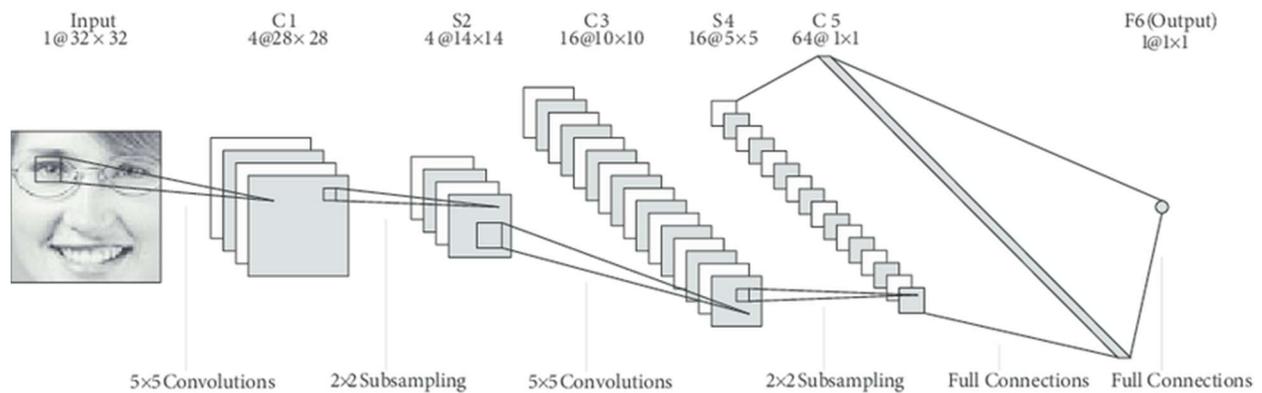

Figure 3. Classical LeNet-5 CNN for face recognition [12]

In Figure 3, it shows how a classical LeNet-5 CNN works for face recognition. The network was proposed by LeCun et al [13], and it is composed by below layers:

Convolution layer: The convolutional layer simulates the process of extracting some primary visual features by using simple methods of local connection and weight sharing to simulate simple cells with local receptive fields. Local connection means that each neuron on the convolutional layer is connected with the neurons in the fixed area in the previous feature map;

weight sharing means that the neurons in the same feature map use the same connection strength and the previous layer. Connection, can reduce the network training parameters, the same set of connection strength is a feature extractor, which is realized as a convolution kernel in the process of calculation, and the convolution kernel value is randomly initialized first, and finally determined by network training.

The pooling/sampling layer: The pooled layer simulates complex cells as a process of screening and combining primary visual features into more advanced, abstract visual features. It is implemented by sampling in the network. After sampling by the pooling layer, the number of output feature maps is unchanged, but the size of the feature map becomes smaller, which has the effect of reducing the computational complexity and resisting small displacement changes.

The pooling layer proposed in this paper adopts large-value sampling, and the sampling size is 2*2, that is, the input feature map is divided into non-overlapping 2*2 rectangles, and the maximum value is taken for each rectangle, so the output feature map is output. Both the length and the width are half of the input feature map. The neurons in the pooled layer defined in this paper do not have the learning function.

The fully connected layer: In order to enhance the nonlinearity of the network and limit the size of the network, the network extracts features from the four feature extraction layers and accesses a fully connected layer. Each neuron of the layer is interconnected with all neurons of the previous layer. The same layer of neurons are not connected.

## 3.3.2 BUILD SIAMESE NETWORK WITH CNN

After comparing different neural networks and their characteristics, we used Siamese network to resolve the problem. The Siamese network is neural network for measuring of similarity. It can be used for category identification, classification, etc., in the scenario when there are many categories, but the number of samples per category is small. The traditional classification method for distinguishing is to know exactly which class each sample belongs to and need to have an exact label for each sample. And the relative number of tags is not too much. These methods are less applicable when the number of categories is too large and the number of samples per category is relatively small. In fact, it is also very well understood. For the entire data set, our data volume is available, but for each category, there can be only a few samples, then using the classification algorithm to do it, because each category of samples is too Less, we can't train any good results at all, so we can only find a new way to train this data set, thus proposing the Siamese network, as showing in Figure 4.

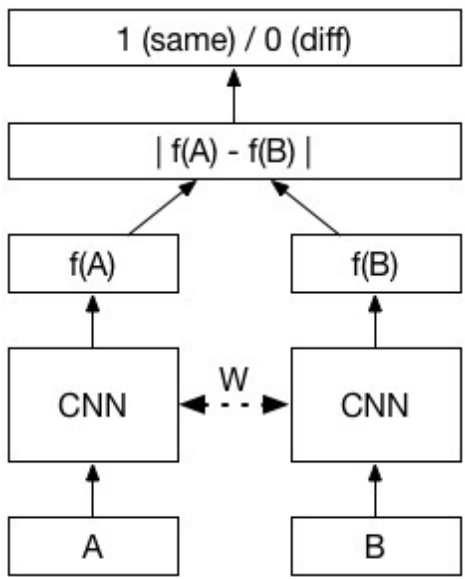

Figure 4. Siamese Network Work Flow [14]

The Siamese network learns a similarity measure from the data and uses the learned metric to compare and match the samples of the new unknown category. This method can be applied to classification problems where the number of classes is large, or the entire training sample cannot be used for previous method training.

The machine we used for this article is on Ubuntu 18 operating system. The CPU is Intel(R) Core(TM) i5-7300HQ CPU @ 2.50GHz with 4 cores. The memory is dual channel DDR 4 8GB SDRAM. We train our neural network using NVIDIA GPU GeForce GTX 1050 with 4GB GPU RAM. The GPU driver version is 390.48. We use Compute Unified Device Architecture (CUDA) version 9.0, and NVIDIA CUDA Deep Neural Network (cuDNN) version 7.0 for CUDA 9.0. We build the Siamese network with below parameters:

```
SiameseNetwork(
  (cnn1): Sequential(
    (0): ReflectionPad2d((1, 1, 1, 1))
    (1): Conv2d(1, 4, kernel_size=(3, 3), stride=(1, 1))
    (2): ReLU(inplace)
    (3): BatchNorm2d(4, eps=1e-05, momentum=0.1, affine=True, track_
running_stats=True)
    (4): ReflectionPad2d((1, 1, 1, 1))
    (5): Conv2d(4, 8, kernel_size=(3, 3), stride=(1, 1))
    (6): ReLU(inplace)
    (7): BatchNorm2d(8, eps=1e-05, momentum=0.1, affine=True, track_
running_stats=True)
    (8): ReflectionPad2d((1, 1, 1, 1))
    (9): Conv2d(8, 8, kernel_size=(3, 3), stride=(1, 1))
    (10): ReLU(inplace)
    (11): BatchNorm2d(8, eps=1e-05, momentum=0.1, affine=True, track
_running_stats=True)
  )
  (fc1): Sequential(
    (0): Linear(in_features=80000, out_features=500, bias=True)
    (1): ReLU(inplace)
    (2): Linear(in_features=500, out_features=500, bias=True)
    (3): ReLU(inplace)
    (4): Linear(in_features=500, out_features=5, bias=True)
  )
```

)

### 3.3.3 TRAIN THE NEURAL NETWORK

The face database we choose is ORL [15]. The ORL face database consists of 400 pictures of 40 people, that is, 10 pictures per person. The face has expressions, tiny gestures and so on. The training processing is performed on the two databases, and 90% of the faces in the library are randomly selected as the training set, and the remaining 10% of the faces are used as test sets, and then the faces in the two sets are standardized. The training process was using GPU, as shown in Figure 5 and Figure 6. We can see during training, the CPU usage went to 100%, and the working temperature increased dramatically.

```
+-----------------------------------------------------------------------------+
| NVIDIA-SMI 390.48                 Driver Version: 390.48                     |
|-------------------------------+----------------------+----------------------+
| GPU  Name        Persistence-M| Bus-Id        Disp.A | Volatile Uncorr. ECC |
| Fan  Temp  Perf  Pwr:Usage/Cap|         Memory-Usage | GPU-Util  Compute M. |
|===============================+======================+======================|
|   0  GeForce GTX 1050     Off | 00000000:01:00.0 Off |                  N/A |
| N/A   35C    P8    N/A /  N/A |    147MiB /  4042MiB |      0%      Default |
+-------------------------------+----------------------+----------------------+

+-----------------------------------------------------------------------------+
| Processes:                                                       GPU Memory |
|  GPU       PID   Type   Process name                             Usage      |
|=============================================================================|
|    0      1037      G   /usr/lib/xorg/Xorg                           78MiB |
|    0      1223      G   /usr/bin/gnome-shell                         68MiB |
+-----------------------------------------------------------------------------+
```

Figure 5. GPU Usage before training.

Figure 6. GPU usage during training.

We trained the model with 100 epochs. A sample training loss is showed below, we can see the training loss goes down significantly during the early epochs, and converged to 0.0067 at last. Figure 8 shows clearly how the trend of training loss goes down as the epoch increases.

```
Epoch number 0
 Current loss 2.055403470993042

Epoch number 1
 Current loss 0.9863696694374084

Epoch number 2
 Current loss 0.8137061595916748

…
…
…

Epoch number 6
 Current loss 0.5110232830047607

Epoch number 7
 Current loss 0.36760351061820984
```

```
Epoch number 8
 Current loss 0.346961110830307

Epoch number 9
 Current loss 0.23092612624168396
…
…
…

Epoch number 33
 Current loss 0.019308971241116524

Epoch number 34
 Current loss 0.02861899696290493

Epoch number 35
 Current loss 0.047534599900245667

…
…
…

Epoch number 70
 Current loss 0.027325695380568504

Epoch number 71
 Current loss 0.019941458478569984

…
…
…

Epoch number 97
 Current loss 0.00932073313742876

Epoch number 98
 Current loss 0.00708159850910306

Epoch number 99
 Current loss 0.006693363655358553
```

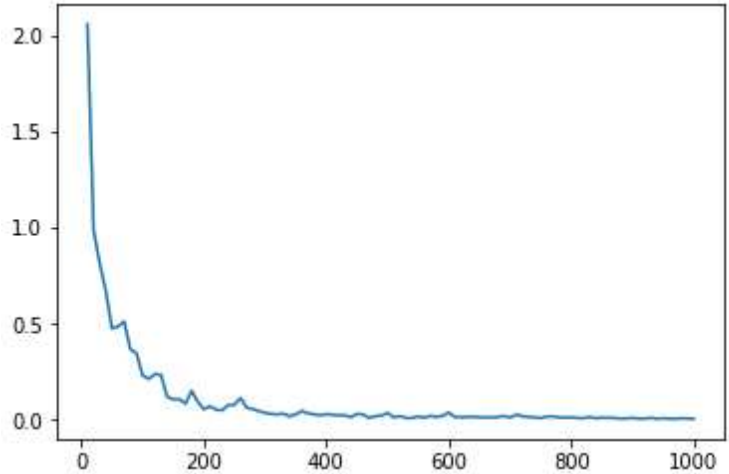

Figure 8. Training loss vs. Epoch

### 3.3.4 MODEL VERIFICATION

The input of the neural network is an image of human face, and the output of the neural network is a vector of 5 dimensions. A sample output looks like below:

```
Vector of Face 1: Variable containing:
 1.7350  0.2165  1.0214  1.5764  2.2253
[torch.cuda.FloatTensor of size 1x5 (GPU 0)]
```

To calculate if the faces on two images come from the same person, we need to calculate the similarity of the two images, aka, the Euclidean distance between two vectors. Below is an example output of different people identified by our model, and the images are shown in Figure 9:

```
Vector of Face 1: Variable containing:
 1.7350  0.2165  1.0214  1.5764  2.2253
[torch.cuda.FloatTensor of size 1x5 (GPU 0)]
Vector of Face 2: Variable containing:
-0.7570  1.5081  0.3380  1.5524 -0.0977
```

```
[torch.cuda.FloatTensor of size 1x5 (GPU 0)]

Distance between Face1 Vector and Face2 Vector: Variable containing:

 3.7070

[torch.cuda.FloatTensor of size 1x1 (GPU 0)]
```

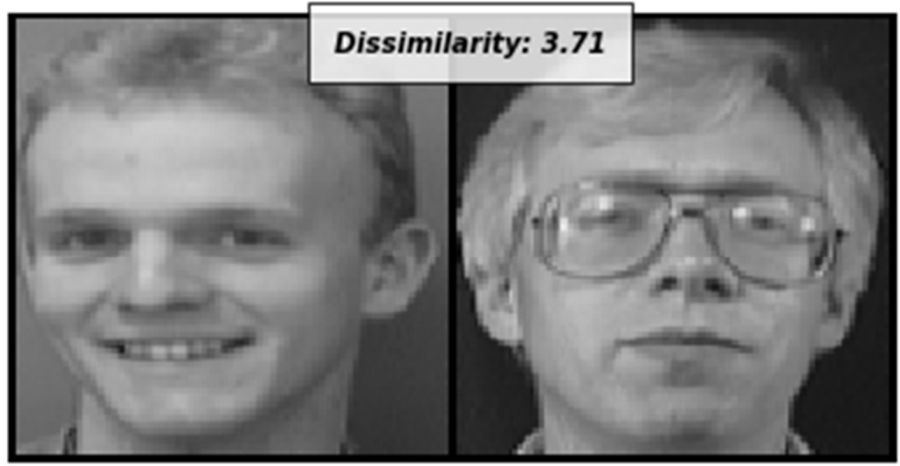

Figure 9. Different people with high Euclidean distance

Below is an example output of same person with different pose, identified by our model, and the images are shown in Figure 10:

```
Vector of Face 1: Variable containing:

 1.7350  0.2165  1.0214  1.5764  2.2253

[torch.cuda.FloatTensor of size 1x5 (GPU 0)]

Vector of Face 2: Variable containing:

 1.6301  0.7585  1.1658  1.6345  2.2486

[torch.cuda.FloatTensor of size 1x5 (GPU 0)]

Distance between Face1 Vector and Face2 Vector: Variable containing:

 0.5741

[torch.cuda.FloatTensor of size 1x1 (GPU 0)]
```

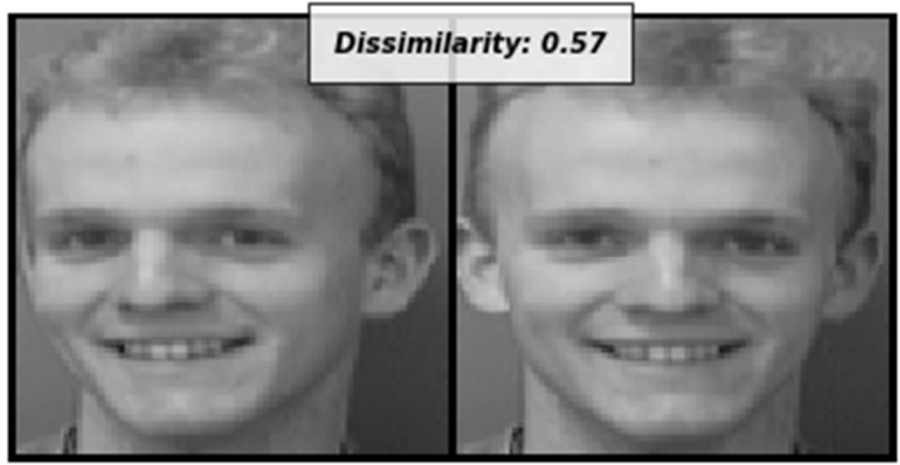

Figure 10. Same person, even with different pose, has small Euclidean distance

# Chapter 4

# Building Robust Face Recognition System

## 4.1 INTRODUCTION

High accuracy face recognition models have been reported in scientific researches by giant technology companies and research institutions, as shown in Figure 11. But all these ground-breaking result still stays in the laboratory. Applications of face recognition in the real world is hard to see. For example, you cannot see it be used in DMV to void fake id, you cannot see it in a daycare to make sure the right person pick up the right kid, you cannot see it in a fitness club to make customer check-in more pleasant.

| Method | Net. Loss | Outside data | # models | Aligned | Verif. metric | Layers | Accu. |
|---|---|---|---|---|---|---|---|
| DeepFace [97] | ident. | 4M | 4 | 3D | wt. chi-sq. | 8 | 97.35±0.25 |
| Canon. view CNN [115] | ident. | 203K | 60 | 2D | Jt. Bayes | 7 | 96.45±0.25 |
| DeepID [92] | ident. | 203K | 60 | 2D | Jt. Bayes | 7 | 97.45±0.26 |
| DeepID2 [88] | ident. + verif. | 203K | 25 | 2D | Jt. Bayes | 7 | 99.15±0.13 |
| DeepID2+ [93] | ident. + verif. | 290K | 25 | 2D | Jt. Bayes | 7 | 99.47±0.12 |
| DeepID3 [89] | ident. + verif. | 290K | 25 | 2D | Jt. Bayes | 10-15 | 99.53±0.10 |
| Face++ [113] | ident. | 5M | 1 | 2D | L2 | 10 | 99.50±0.36 |
| FaceNet [82] | verif. (triplet) | 260M | 1 | no | L2 | 22 | 99.60±0.09 |
| Tencent [8] | - | 1M | 20 | yes | Jt. Bayes | 12 | 99.65±0.25 |

Figure 11. Face recognition models and their accuracy [16].

The key gap between face recognition research and industrial usage is the application. You can see different algorithm companies provided different APIs, but how to turn the API into the real product is a tough problem laying in front of potential industrial users.

## 4.2 SYSTEM ARCHITECTURE

Many aspects need to be take into consideration when building towards a commercial usable system. We are targeting at a system architecture that is high performance, scalable, agile, and low cost. The high performance means the system will give out result at milliseconds level, and have high threshold with high concurrency. Scalable means the system can scale well as the needs increases, from a single node machine to a multi node cluster. Agile means the system should be easily modifiable, and be able to apply to different domain with ease. Low cost not only means the deployment cost is low, also the development cast and maintenance cost. To achieve all the goals, we need to resolve many practical problems.

### 4.2.1 CHOOSE BETWEEN CPU AND GPU

Although GPU was not designed for neural network initially, but it has the features which put it into a better position than CPU for the neural network calculations: 1. Provides the infrastructure of multi-core parallel computing, and has a large number of cores, which can support parallel computing of large amounts of data. Parallel computing or parallel computing is relative to serial computing. It is an algorithm that can execute multiple instructions at a time, with the goal of increasing the speed of calculations and solving large and complex computational problems by expanding the problem solving scale. 2. Has a higher speed of memory access. 3. Has higher floating point computing power. Floating-point computing power is an important indicator of multimedia and 3D graphics processing related to processors. In today's computer technology, due to the application of a large number of multimedia technologies, the calculation of floating point numbers has been greatly increased, such as the rendering of 3D graphics, so the ability of floating point computing is an important indicator to examine the computing power of the processor.

A test done by V Chu [17] shows that, depending on the GPU you choose, the performance gain of a GPU over CPU on neural network calculation can be between 43 times to 167 times, as shown in Figure 12. So, to achieve the high performance, our system will need to use GPU for the neural network related processes.

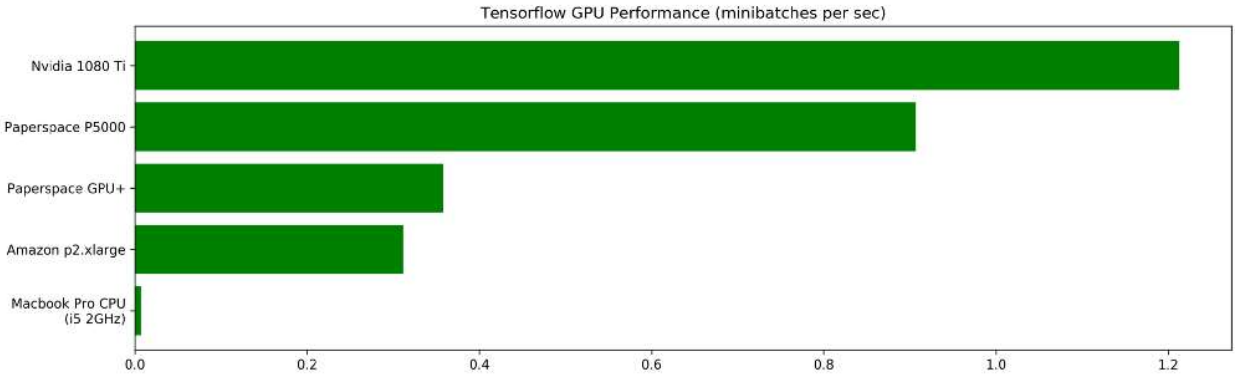

Figure 12, Tensorflow Performance comparison between GPUs and CPU [17]

## 4.2.2 CHOOSE BETWEEN ALL-IN-ONE AND CLIENT-SERVER ARCHITECTURE

The demos from many face recognition research institutions are using all-in-one architecture. In these examples [18], all processing unit are inside same system. This is very good for prototyping, but has issues on industrial applications. The first problem to be considered is the size, cost and power consumption of a graphic card. Figure 13 shows a RTX 2080 Ti graphic card held by NVIDIA CEO Jensen Huang. The size is so big that it cannot be put into small size appliances. The power consumption is 280W, with huge amount of heats generated when running at full potential. The price is no less than $999 for 1 pieces, which limit bulk deployment of it.

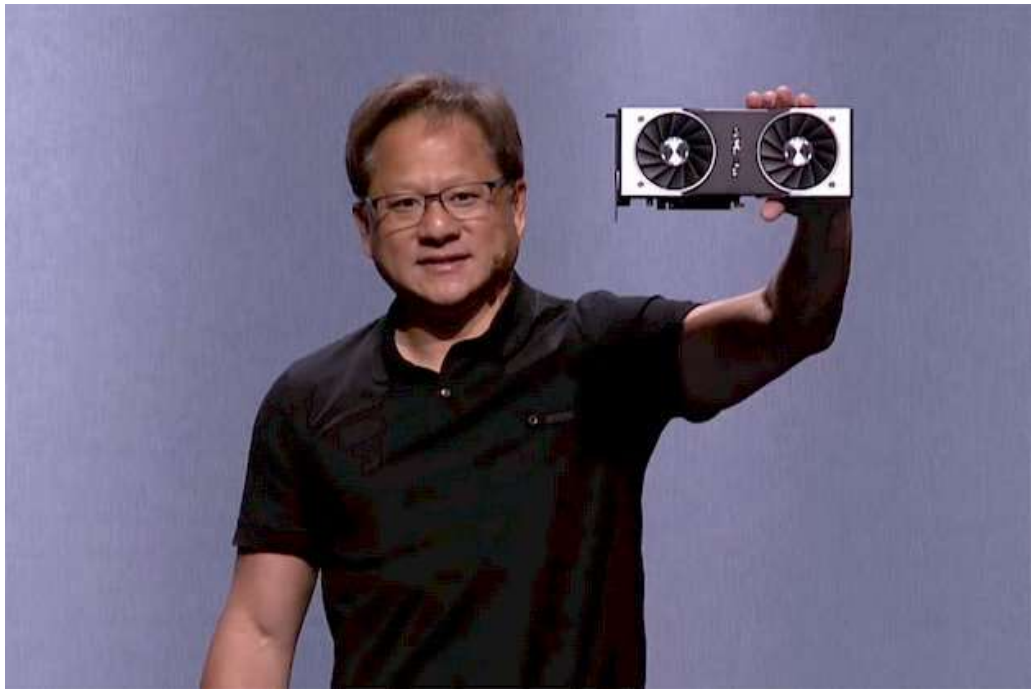

Figure 13. A RTX 2080 Ti graphic card held by NVIDIA CEO Jensen Huang [19]

Despite all the limitations of the graphic card, its computational power is superior. In our stress test with a GTX 1080 Ti graphic card, it can extract facial features from 200 pictures at same time, with an average of 25 milliseconds processing time per picture. All the limitations and features are implicating that the server-client architecture will perform well.

### 4.2.3 THE CLIENT-SERVER ARCHITECTURE WITH GPU

Figure 14 shows the client-server architecture with GPU we designed to fulfill the requirement. The whole system has 2 sub-systems: user registration system and real time recognition system, and these 2 sub-systems share compute-heavy components to reduce the complexity and cost. The client side of user registration system will take user picture and user input, i.e. employee id. Then the picture and user input is sent to the server, where GPU resides. The GPU will process the picture, first detect human face and then extract the picture into vector. The vector is then stored into database with the user input.

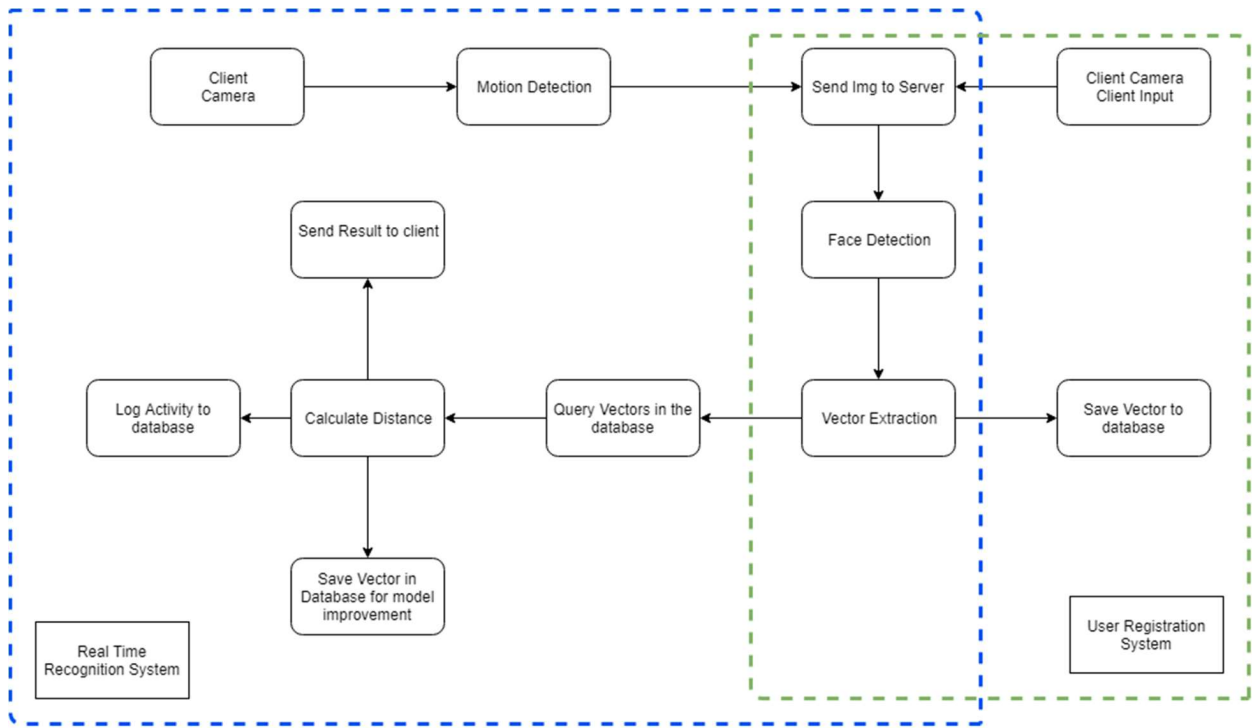

Figure 14. Client-server architecture with GPU

In the real time recognition system, the client camera will be used to capture real time videos. In our first design, the client will send video feed to the server no matter what was captured. During the stress test, we found this method will put too much load into the server. So we added a motion detection component to the client. This component can be done with opencv libraries, and can be deployed on both desk-top level clients, and mobile clients. The motion detection will significantly reduce the amount of image which need to be processed for face detection.

After a motion is detected, the video feed will be processed and only send n frames of pictures to the server. The server's face detection components will try to detect the face from the image, it will be able to detect multiple faces from single image. If there is no face detected, the image will be discarded. If one or multiple faces were detected, each face will be cropped from the original image, aligned, resized, and marked with a unique ID and send to the vector extraction processer. The vector extraction processer will read the pre-processed human face image and extract each face into a vector. The vector will be combined with the unique request ID, and start a query into the face feature database. The Euclidean distance will be calculated against the current face vector and the face vectors retrieved from the database. This step can be done by CPU. After all the distances have been calculated, they will be sorted and generated the list n distances. A message similar to below format will be generated:

```
{
        request_id:<x>,
        {
                user_id[1]:<x>
                distance:<x>
        },
        {
                user_id[2]:<x>
                distance:<x>
        },
        {
                user_id[3]:<x>
                distance:<x>
        }
}
```

This message will be sent back to the client, and on the client, the customer can have their own rules on what will happen, i.e. allow employee badge-in, tell if an I.D. card is fake and etc. In the meanwhile, the message can be sent to a different database, which logs the activities, and it can hook up with customers' ERP system. The images can also be saved for further training, to improve the model performance on this particular group of people.

## 4.2.4 DATABASE DESIGN

To preserve data, this system will use its own database. We will use MySql relational database management system as an example in this article. An obvious way to for storing the human face picture into the database is to use blob (binary large object). But if the image are stored in the database, every time to do the face compare, the GPU need to re-extract the vector from the images. For large dataset, the wait will be long. To avoid this issues, we will first extract the vectors from human face image, and only save the vectors into the database. And we also need to associate the vector with an identification property. User_id/employee_id/customer_id will be good choice for this identification property. So, depending on how many vectors will the model extract, the main table structure in the database will look like below:

| ID | Vector1 | Vector2 | Vector3 | … | … | … | Vector<n-1> | Vector<n> |
|----|---------|---------|---------|---|---|---|-------------|-----------|
|    |         |         |         |   |   |   |             |           |

According to database normalization rules, the database belongs to this system shouldn't store anything other than the ID and face vector. Since many user information can change rapidly inside on organization, bring any other column into this database will be redundant and require additional work to maintain data integrity. The database can retrieve/send information to other enterprise database like ERP, Clarify and etc. with just ID column, like showing in Figure 15.

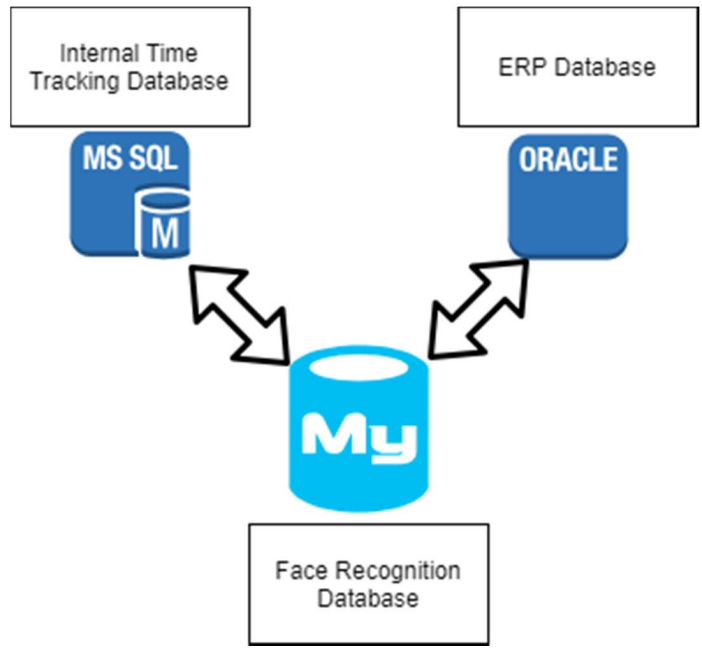

Figure 15. Database linking to each other.

According to our experiment, one tweak in the database design is to not make the ID column a unique key or primary key column. By doing this, the performance can be significantly improved. For example, after the system is online for 10 days, you gathered one face with ID=9 10 times, with slightly different vectors every time. You can save all these 10 vectors into this main table. One the 11th day, when person with ID=10 go through the camera, the picture is processed, and the top 3 candidates from the processing engine will show ID=10 more than 1 times. This result will give client more confidence to justify the person is the one with ID=10.

### 4.2.5 NODE FRAMEWORK AS SERVER

Node.js is a JavaScript runtime environment, released in May 2009 by Ryan Dahl, which essentially encapsulates the Chrome V8 engine. Node.js is neither a JavaScript framework, nor a browser-side library. Node.js is a development platform that lets JavaScript run on the server side, making JavaScript a scripting language that is on par with server-side languages like PHP, Python, Perl, and Ruby.

The V8 engine itself uses some of the latest compilation techniques. This allows code written in a scripting language such as JavaScript to run at a much faster speed and saves development costs. The demand for performance is a key factor in Node. JavaScript is an event-driven language, and Node takes advantage of this to write highly scalable servers. Node uses an architecture called an "event loop" that makes writing a highly scalable server easy and secure. There are many different techniques for improving server performance. Node chose an architecture that improves performance while reducing development complexity. This is a very important feature. Concurrent programming is often complex and full of mines. Node bypasses these, but still provides good performance.

Node uses a series of "non-blocking" libraries to support the way the event loops. Essentially, it provides an interface for resources such as file systems and databases. When a request is sent to the file system, there is no need to wait for the hard disk (addressing and retrieving the file), and the non-blocking interface notifies Node when the hard disk is ready. The model simplifies access to slow resources in an extensible way, intuitive and easy to understand. Especially for users who are familiar with DOM events such as onmouseover and onclick, there is a feeling of deja vu.

Although running JavaScript on the server side is not unique to Node, it is a powerful feature. I have to admit that the browser environment limits our freedom to choose a programming language. The desire to share code between any server and an increasingly complex browser client application can only be achieved through JavaScript. Although there are other platforms that support JavaScript running on the server side, because of the above characteristics, Node has developed rapidly and become top choice for many developers.

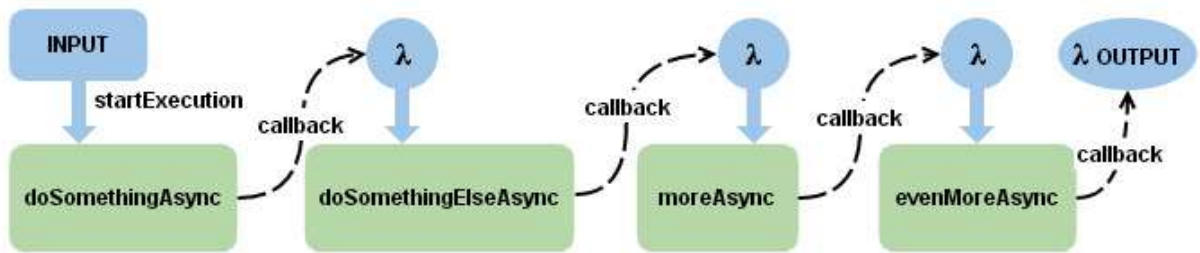

Figure 16. Callback in Node.js [20]

Figure 16 is a diagram of how callback works in Node. Although in the face recognition system, the I/O is not the bottle neck, but Node still is our best choice. Because for GPU to detect face and extract vector from face image, and average of 25 milliseconds will be consumed. With the non-block feature of Node, the whole process will not just wait there for 25 seconds each time a request comes. The Node sever will send the request to GPU, then keeps handling new requests. After GPU finished its work, it will run the call back function, and the server will pick up what's left on the request and move one. This increased the concurrency of the system significantly.

Figure 17 shows how the node server react with the neural network model in our system. When request comes in, the node sever will send the request to the neural network model. The neural network, which can be deployed in any format, will take the task and run it on GPU. While this workload is running, node server will not wait for it to finish, but keeps accepting request from the caller. After the neural network model retuned the calculation result, node server will package it with other necessary information, and send the feedback to the caller.

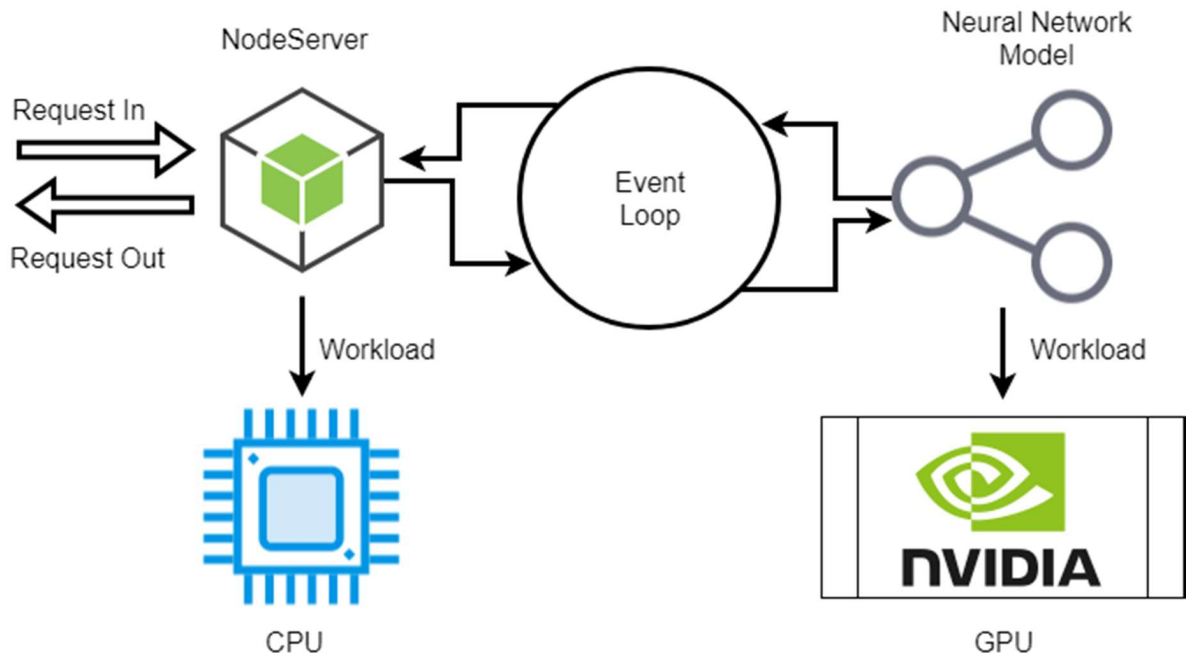

Figure 17. Node Server work with Neural Network Model.

Below is our Node Server deployment with quest queueing:

```
var blockList = [];
var waitList = [];
var list1 = [];  //test list

app.post('/postUid', function (req, res) {
  var uid1 = req.body.uid1;
  var uid2 = req.body.uid2;
  var uid3 = req.body.uid3;

  var value1 = req.body.value1;
```

```javascript
  var value2 = req.body.value2;
  var value3 = req.body.value3;

  function timeOutBlockList(listElement){
    setTimeout(function(){
        var listPos = blockList.indexOf(listElement);
        console.log(new Date(),'BlockList: To be removed From block list:',li
stElement);
        // blockList[listElement] = 'False';    // to set value to this eleme
nt.
        blockList.splice(listPos,1);   // to remove this element.
        //console.log(new Date(),'New block list',blockList);
    }, blockTime);
  }

  function timeOutDisplayList(listPos,uid,waitTime){
    setTimeout(function(){
        console.log(new Date(),'DisplayList: To be emptyed: position: ', list
Pos);
        if (displayList[listPos].uid === uid ) {
          if (displayList[listPos].stat === 'Replaceable') {
            displayList[listPos].stat = 'Empty';
          }
          request.post({url:'http://localhost:3000/postInfo', form: {slot:di
splayList[listPos].slot,uid:'', imgSrc:displayList[listPos].imgSrc,username:d
isplayList[listPos].username, title:displayList[listPos].title}}, function (e
rr,response, body) {
            if (err) {
              return console.error(err);
            } else {
              console.log(new Date(),"Timeout DL, post to SSE sucesss",listPo
s,uid);
            }
          });
              // to set value to this element.
          //displayList.splice(listPos,1);   // to remove this element.
          //console.log(new Date(),'New display list',displayList);

        } else {
          console.log(new Date(),uid," no longer on the display list")
        }
    }, waitTime );
  }
```

```
  function addToDL(listPos,uid){
    displayList[listPos].stat = 'OnScreen';
    displayList[listPos].uid = uid;
    // query the user info with UID
    var sqlStmt = 'SELECT uid, username, title from client_users where uid =
\"' + uid + '\";';
    //console.log(new Date(),sqlStmt);
    connection.query(sqlStmt, function (err, rows) {
      if (err) {
        // throw err;
        console.log(new Date(),err);
      }
      else {
        if (rows.length > 0) {
          console.log(new Date(),'DB query result: ', rows[0].uid, rows[0].us
ername, rows[0].title);
          displayList[i] = {
            uid:rows[0].uid,
            slot : i+1,
            imgSrc:'userimg/'+rows[0].uid + '.jpg',
            username:rows[0].username,
            title:rows[0].title
          };

        request.post({url:'http://localhost:3000/postUinfo', form: {slot:di
splayList[i].slot,uid:displayList[i].uid, imgSrc:displayList[i].imgSrc,userna
me:displayList[i].username, title:displayList[i].title}}, function (err,respo
nse, body) {
            if (err) {
              return console.error(err);
            } else {
              console.log(new Date(),"Add to DL: Post to SSE sucesss",listPo
s,uid);
            }
          });
          // below is the one without handle exception where sse is down.
          //request.post('http://localhost:3000/postUinfo').form({slot:displa
yList[i].slot,uid:displayList[i].uid, imgSrc:displayList[i].imgSrc,username:d
isplayList[i].username, title:displayList[i].title});

        } else {
          console.log(new Date(),'User not in database: ', uid1);
```

```
        }
      }
    });

    oneSecFunc(listPos,uid);
  }

  function oneSecFunc(listPos,uid) {
    setTimeout(function(){
      if (waitList.length > 0) {
        var wl1 = waitList.shift();
        console.log(new Date(),listPos, " is picking from waitlist for ",wl
1);
        // play out animation, about 500
        displayList[listPos].stat = 'pickedUpfromWL';
        timeOutDisplayList(listPos,uid,0);
        //-- wrong --setTimeout(addToDL(listPos,wl1),500);
        console.log(new Date(),wl1,' will be added to DL');
        setTimeout(function (){addToDL(listPos,wl1);},500);
      } else {
        displayList[listPos].stat = 'Replaceable';
        timeOutDisplayList(listPos,uid,displayTime - 1000);
      }
    } , 1000);
  }

  // blockList.push(uid1); //test
  if (value1 > yz1) {
    if (blockList.includes(uid1)) {
      console.log(new Date(),"Uid1 already in BlockList. Uid1 is: ", uid1);
    } else {
      console.log(new Date(),"Uid1 not in BlockList. Uid1 is: ", uid1);

      // add this user to block list
      blockList.push(uid1);
      timeOutBlockList(uid1);

      // add this user to display list
      var displayed = false;
      for (var i=0; i<displayList.length; i++) {
        if (displayList[i].stat === 'Empty') {
          console.log(new Date(),"Dispaly on Display list index: ",i);
```

```
            addToDL(i,uid1);
            displayed = true;
            break;
          }
        }

        // Add a waitDisplayList to handle displayed = false
        console.log(new Date(),'Displayed?: ', displayed);
        if ( displayed == false ) {
          var pushed = false;
          for (var i=0; i<displayList.length; i++) {
            if (displayList[i].stat === 'Replaceable') {
              console.log(new Date(),"Push to Display list index: ",i);
              timeOutDisplayList(i,displayList[i].uid,0);
              displayList[i].stat = "waitForPush"
              setTimeout(function (){addToDL(i,uid1);},500);
              pushed = true;
              break;
            }
          }
          console.log(new Date(),'Pushed?: ', pushed);
          if ( pushed == false ) {
            waitList.push(uid1)
            console.log(new Date(),'Current wait list',waitList);
          }
        }
        //console.log(new Date(),displayList)
      }
  } else {
    console.log(new Date(),"Largest value lower than ",yz1);
  }

  res.send('Got a POST request:: ' + JSON.stringify(req.body) );
})

app.get('/getUinfo', (req, res) => res.send(JSON.stringify(displayList)))
```

# Chapter 5

# Conclusion and Future Work

## 5.1 CONCLUSION

We proposed to build a high performance, scalable, agile, and low cost face recognition system. We divide the proposed approach into several small sub projects. First, we studied neural network and convolutional neural network. Based on the theory of deep learning, we built the Siamese network which will train the neural network based on similarities. Then we examine and compare the available open source data set, we chose ORL dataset and trained the model with GPU. The model will take a human face image and extract it into a vector. Then the distance between vectors are compared to determine if two faces on different picture belongs to the same person.

Then we did the study, compare, design and build a system to work with the neural network model. The system uses client-server architecture. GPU is used on the server side to provide high performance. We also de-coupled the main components of the system to make it flexible and scalable. We used the non-block and asynchronies features of Node.JS to increase the system's concurrency. Since the entire system is modularized, it can be used in different domains, thus reduced the development cost.

## 5.2 FUTURE WORK

When build the neural network model, there are many parameters which can be tuned to increase the model performance. We can keep tuning our models to increase its accuracy.

Also, for a trained base model, we can re-train it using a specific dataset. So another way to increase the whole system's performance is to capture the specific people's images and re-train the model based on this small dataset. For example, if an organization with 3000 people uses this system, the model can be trained to be very accurate on these 3000 people. We can employ and automate this feature into the system.